\documentclass[lettersize,journal]{IEEEtran}
\usepackage{amsmath,amsfonts}
\usepackage{array}
\usepackage[caption=false,font=normalsize,labelfont=sf,textfont=sf]{subfig}
\usepackage{textcomp}
\usepackage{stfloats}
\usepackage{url}
\usepackage{verbatim}
\usepackage{graphicx}
\usepackage{cite}
\usepackage{multirow}
\usepackage[table]{xcolor}
\usepackage[normalem]{ulem}
\useunder{\uline}{\ul}{}
\usepackage[linesnumbered,ruled,vlined]{algorithm2e}
\usepackage[colorlinks, linkcolor=blue, anchorcolor=blue, citecolor=blue]{hyperref}

\begin{document}

\title{AW-MoE: All-Weather Mixture of Experts for \\ Robust Multi-Modal 3D Object Detection}

\author{Hongwei~Lin,
        Xun~Huang,
Chenglu~Wen$^\dag$,~\IEEEmembership{Senior~Member,~IEEE},
and~Cheng~Wang,~\IEEEmembership{Senior~Member,~IEEE}
        
\thanks{Hongwei~Lin, Xun~Huang, Chenglu~Wen, and Cheng~Wang are with the Fujian Key Laboratory of Urban Intelligent Sensing and Computing and the School of Informatics, Xiamen University, Xiamen, FJ 361005, China. (E-mail: \{greatlin, huangxun\}@stu.xmu.edu.cn; \{clwen, cwang\}@xmu.edu.cn). Xun~Huang is also with Zhongguancun Academy, Beijing, China.}
\thanks{$^\dag$ Corresponding author.}}



\maketitle

\begin{abstract}
Robust 3D object detection under adverse weather conditions is crucial for autonomous driving. However, most existing methods simply combine all weather samples for training while overlooking data distribution discrepancies across different weather scenarios, leading to performance conflicts. To address this issue, we introduce AW-MoE, the framework that innovatively integrates Mixture of Experts (MoE) into weather-robust multi-modal 3D object detection approaches. AW-MoE incorporates \textbf{I}mage-guided \textbf{W}eather-aware \textbf{R}outing (IWR), which leverages the superior discriminability of image features across weather conditions and their invariance to scene variations for precise weather classification. Based on this accurate classification, IWR selects the top-K most relevant \textbf{W}eather-\textbf{S}pecific \textbf{E}xperts (WSE) that handle data discrepancies, ensuring optimal detection under all weather conditions. Additionally, we propose a \textbf{U}nified \textbf{D}ual-\textbf{M}odal \textbf{A}ugmentation (UDMA) for synchronous LiDAR and 4D Radar dual-modal data augmentation while preserving the realism of scenes. Extensive experiments on the real-world dataset demonstrate that AW-MoE achieves $\sim$15\% improvement in adverse-weather performance over state-of-the-art methods, while incurring negligible inference overhead. Moreover, integrating AW-MoE into established baseline detectors yields performance improvements surpassing current state-of-the-art methods. These results show the effectiveness and strong scalability of our AW-MoE. We will release the code publicly at https://github.com/windlinsherlock/AW-MoE.
\end{abstract} 


\section{Introduction}
\label{sec:intro}

\IEEEPARstart{T}{hree-Dimensional} object detection, a fundamental task in 3D computer vision, has significantly advanced autonomous driving and other unmanned systems. Most existing methods rely on the stable performance of sensors, such as LiDAR~\cite{pointrcnn, pointpillars, centerpoint, pv-rcnn, tip1} and cameras~\cite{motal, bevfusion}. However, under adverse weather conditions (e.g., rain, fog, snow), sensor performance degrades, leading to weakened system reliability~\cite{analyse}. 

Therefore, recent studies have explored developing robust 3D object detection techniques under adverse weather conditions. These works pursue robustness through two complementary approaches: the construction of simulation-augmented~\cite{analyse, sunshine, fog} or real-world datasets~\cite{k-radar} at the data level, and the development of multi-modal fusion techniques~\cite{l4dr, towards, availability} at the algorithmic level. However, existing methods primarily simply combine all weather samples for training while overlooking the substantial distribution discrepancies across adverse weather conditions, which may lead to performance conflicts across various scenarios. 

\begin{figure*}[t]
  \centering
   \includegraphics[width=0.9\linewidth]{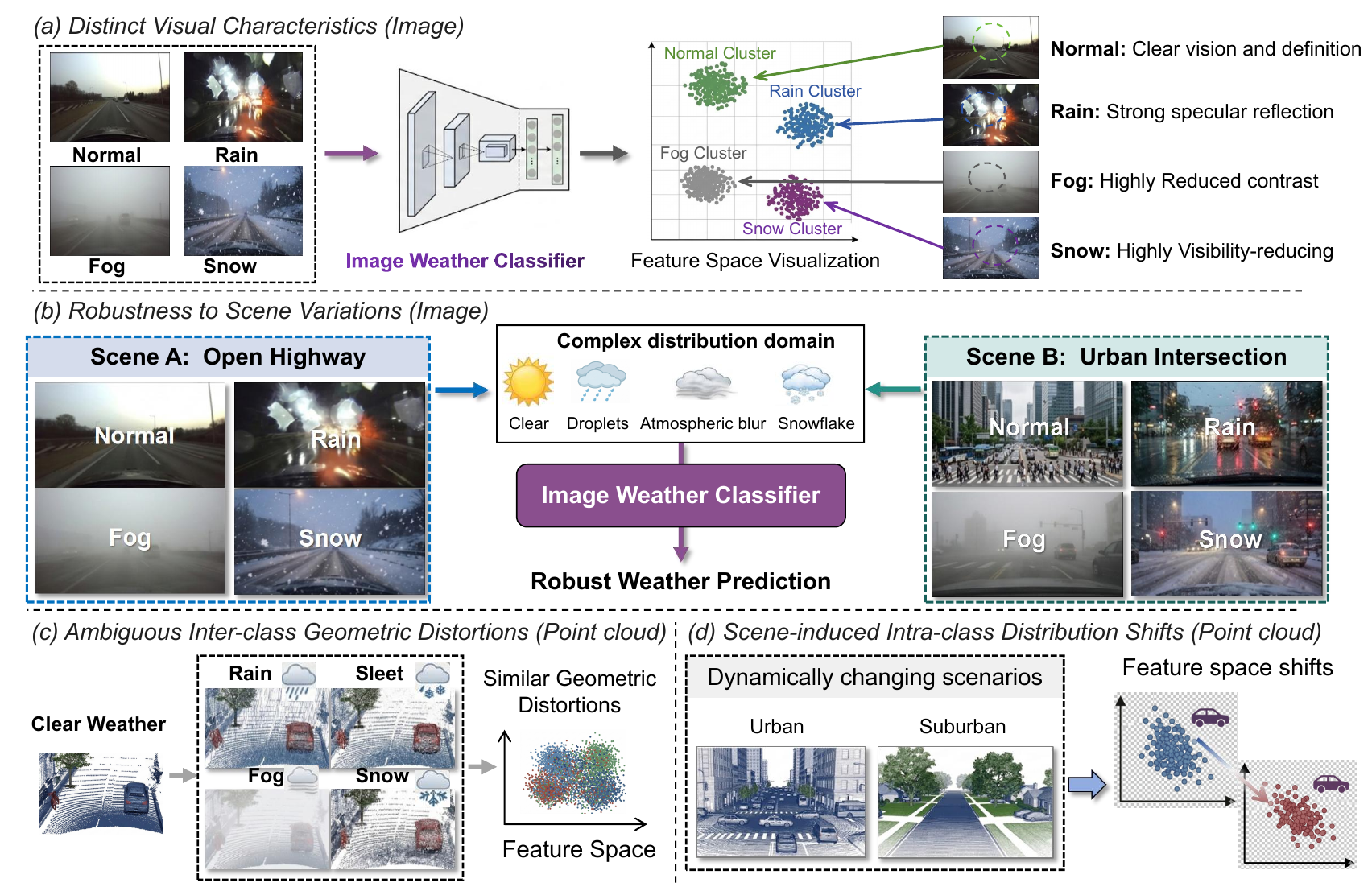}
   \caption{Comparison of weather-type discriminability between camera images and LiDAR point clouds. (a, b) Camera images exhibit distinct visual characteristics and robustness to scene variations, facilitating accurate weather classification. (c, d) In contrast, LiDAR point clouds suffer from ambiguous inter-class geometric distortions and scene-induced intra-class distribution shifts, which obscure the boundaries between different weather conditions.}
   \label{fig:Image Advantage}
\end{figure*}

\begin{figure}[t]
  \centering
   \includegraphics[width=\linewidth]{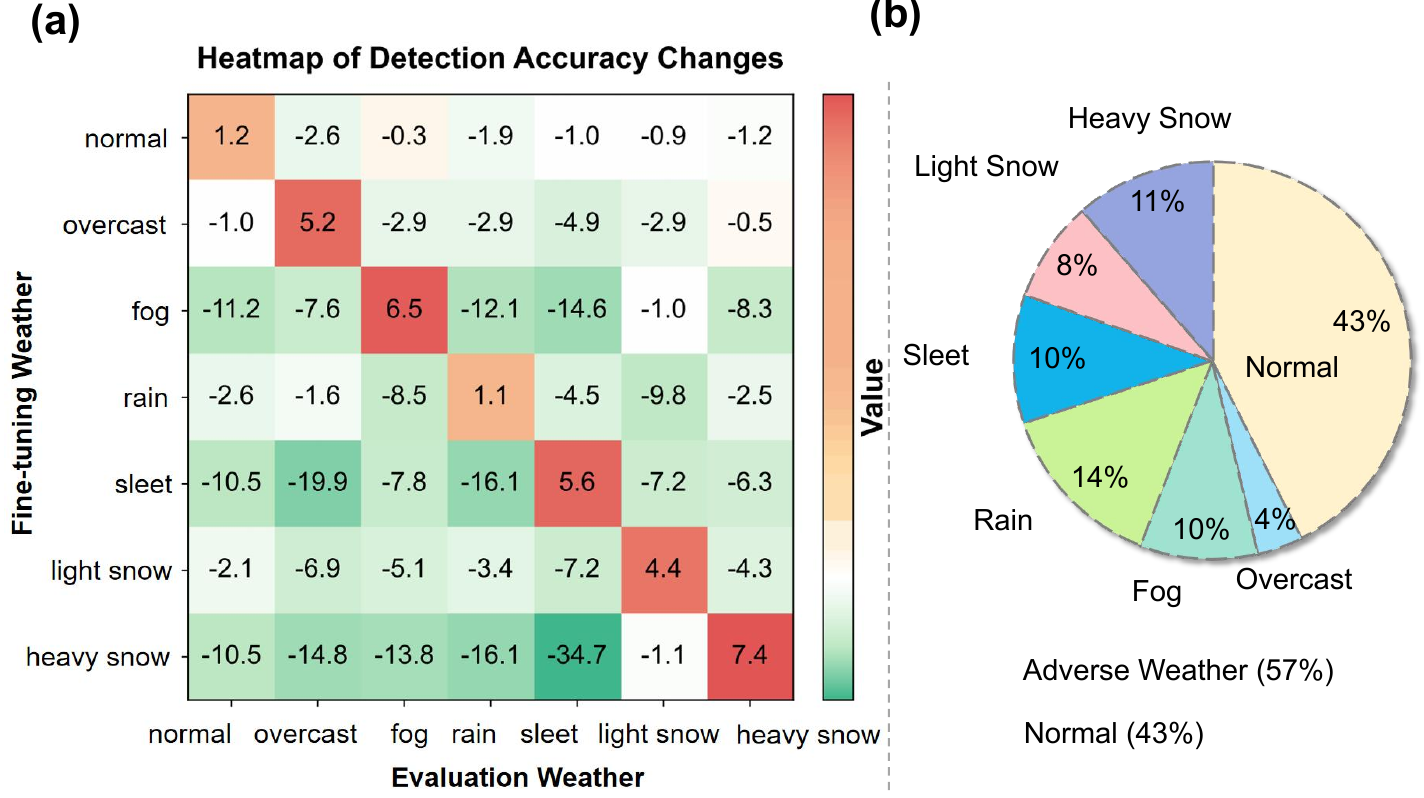}
   \caption{(a) Performance changes of L4DR~\cite{l4dr} after fine-tuning on a single weather condition under different weather scenarios. (b) Statistics of data volume across different weather conditions in the K-Radar dataset~\cite{k-radar}.}
   \label{fig:Data}
\end{figure}

To investigate this, we first explore the influence of weather sample bias through fine-tuning the state-of-the-art L4DR method~\cite{l4dr} separately on each weather subset. As shown in Fig.~\ref{fig:Data} (a), models fine-tuned on a specific weather condition improve in that condition but suffer degraded performance in others. This phenomenon indicates that significant distributional gaps exist across weather conditions, preventing a single model from maintaining optimal performance across all conditions. Moreover, due to the expensive collection of adverse weather data, real-world datasets such as K-Radar~\cite{k-radar} contain far fewer adverse weather samples than normal-weather ones (see Fig.~\ref{fig:Data} (b)). This imbalanced distribution in weather samples tends to bias training toward normal weather conditions, thereby further overlooking adverse weather scenarios.

\IEEEpubidadjcol

To address these challenges, we propose the Adverse-Weather Mixture of Experts (AW-MoE), the first approach that introduces the Mixture of Experts (MoE) technique to enhance the robustness of 3D object detection under adverse weather conditions. AW-MoE leverages the MoE mechanism~\cite{adaptive_moe, sparse_moe} to extend a single-branch detector into a specialized multi-branch architecture, in which each branch is explicitly tailored to a specific weather condition. This design enables robust adaptation to diverse adverse-weather scenarios while incurring negligible inference overhead. 

\begin{figure}[t]
  \centering
   \includegraphics[width=1\linewidth]{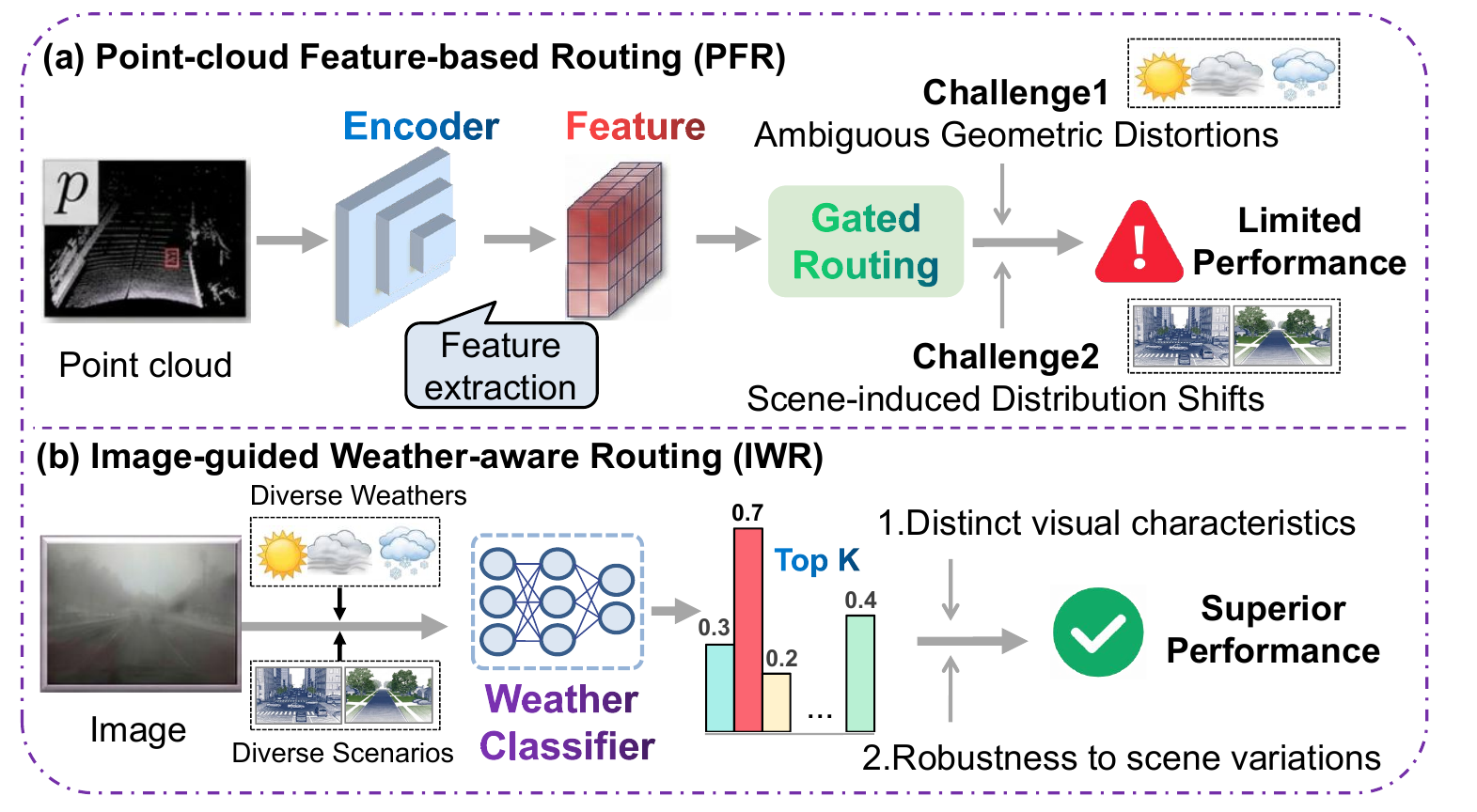}
   \caption{Method comparison between Point-cloud Feature-based Routing (PFR) and the proposed Image-guided Weather-aware Routing (IWR).}
   \label{fig:routing}
\end{figure}

It is worth noting that the effectiveness of Mixture-of-Experts (MoE) in multi-scenario applications heavily relies on optimal expert routing. Standard MoE frameworks~\cite{sparse_moe} typically employ Point-cloud Feature-based Routing (PFR), utilizing input point-cloud features to guide the routing process, as shown in Fig.~\ref{fig:routing} (a). However, PFR exhibits significant limitations in outdoor autonomous driving under adverse weather conditions. First, point clouds suffer from \textit{ambiguous inter-class geometric distortions}, making it difficult to precisely differentiate weather conditions in the feature space (see Fig.~\ref{fig:Image Advantage} (c)). Furthermore, the highly dynamic nature of outdoor environments leads to \textit{scene-induced intra-class distribution shifts}, where point clouds of the same weather exhibit massive variations across different scenes (see Fig.~\ref{fig:Image Advantage} (d)).

In contrast, camera images demonstrate highly favorable properties for weather perception. First, images present \textit{distinct visual characteristics} (see Fig.~\ref{fig:Image Advantage} (a)). For instance, normal weather offers clear vision and high definition, rain introduces windshield droplets and strong specular reflections, and snow presents significant snowflake accumulations. These prominent visual cues enable an Image Weather Classifier to easily distinguish weather conditions in the feature space. Second, images demonstrate strong \textit{robustness to scene variations} (see Fig.~\ref{fig:Image Advantage} (b)). 

Motivated by these observations, we propose an Image-guided Weather-aware Routing (IWR) module, illustrated in Fig.~\ref{fig:routing} (b). IWR leverages an Image Weather Classifier to explicitly identify the weather condition, thereby routing the data to the most suitable weather expert to mitigate data distribution discrepancies. As shown in Table~\ref{tab:Routing}, our IWR achieves a routing accuracy of nearly 99\% across all weather conditions, whereas the baseline PFR struggles significantly to recognize severe weather environments.

Guided by the accurate expert routing of IWR, a Weather-Specific Experts (WSE) module subsequently extracts weather-specific features for the corresponding conditions. Moreover, to mitigate the scarcity of adverse weather samples, we propose a Unified Dual-Modal Augmentation (UDMA) module that performs synchronized data augmentation on both LiDAR and 4D Radar point clouds. Furthermore, we introduce a variant termed AW-MoE-LRC, which integrates image features with the LiDAR and 4D Radar representations. This variant fully exploits the rich semantic information of cameras to achieve enhanced perception performance.

Extensive experiments on the real-world K-Radar~\cite{k-radar} dataset demonstrate that AW-MoE outperforms state-of-the-art methods and shows the extensibility of AW-MoE. Our main contributions are summarized as follows:

\begin{itemize}

\item We propose the Adverse-Weather Mixture of Experts (AW-MoE), which first introduce MoE technique to enhance the robustness of 3D object detection in adverse weather scenarios. The effect is remarkably significant, AW-MoE achieves robust detection performance across all weather conditions.

\item Leveraging the inherent advantages of images in distinguishing weather types, we design the Image-guided Weather-aware Routing (IWR) and Weather-Specific Experts (WSE) modules. This design overcomes the limitations of prior MoE routing approaches under varying weather conditions, thereby enhancing the overall effectiveness of the framework. Additionally, we introduce a tri-modal variant, AW-MoE-LRC, which further incorporates camera features into the LiDAR and 4D Radar modalities.

\item Our AW-MoE is also a highly scalable framework that can be extended to various 3D object detection methods, yielding substantial performance gains under adverse weather conditions. Extensive experiments on real-world datasets demonstrate the superior performance and strong extensibility of our AW-MoE.

\end{itemize}

\section{Related Work}
\label{sec:related}

\subsection{3D Object Detection.}
3D object detection~\cite{std, behind, pb, coin, CPD, smoke, tip2, tip3} is a core task in 3D vision, predominantly relying on raw point clouds like LiDAR. Existing methods are broadly categorized into three types based on data representation: point-based, voxel-based, and point-voxel-based. Point-based methods~\cite{pointrcnn, 3DSSD, Point-gnn} directly sample and aggregate features from raw points. They classify foreground points and predict corresponding 3D bounding boxes. This preserves fine-grained geometric details but incurs high computational costs. Conversely, voxel-based methods~\cite{second, voxel_RCNN, Voxelnet, pointpillars, centerpoint} partition point clouds into regular grids. They aggregate features within each voxel and apply 3D spatial convolutions. Many models~\cite{pointpillars, voxel_RCNN} further compress these features into Bird's Eye View (BEV) space for efficient 2D convolutions, significantly accelerating inference. Point-voxel-based methods~\cite{std, pv-rcnn} integrate both representations to balance efficiency and geometric accuracy. While these approaches achieve impressive accuracy under normal conditions, their performance drops significantly in adverse weather. Environmental interference degrades LiDAR signals, severely compromising the reliability of these conventional methods.

\subsection{3D Object Detection Under Adverse Weather.}
Under adverse weather conditions, the perception capability of sensors such as LiDAR degrades, leading to reduced detection performance~\cite{analyse, performance}. Recent research has extensively explored 3D object detection~\cite{pointpillars, tip4, centerpoint, tip5} under such conditions~\cite{sunshine, robo3d, seeing, analyse, robust_multimodal, l4dr, spg, tip6}. Some works generate simulated adverse weather data (e.g., rain, snow, fog) to train robust detection models~\cite{v2x-r, analyse, sunshine, fog}. In contrast, others focus on real-world datasets such as K-Radar~\cite{k-radar}, which provides multimodal data from LiDAR, 4D Radar, and cameras, and introduces RTNH~\cite{k-radar} using 4D Radar for detection. Furthermore, sensor-fusion methods, including Bi-LRFusion~\cite{Bi-lrfusion}, 3D-LRF~\cite{towards}, and L4DR~\cite{l4dr}, leverage complementary information from LiDAR and Radar to enhance robustness. Although these approaches outperform single-modal methods, they overlook the substantial distribution gaps across different adverse weather conditions. Our experiments reveal that training a single-branch model with mixed-weather data causes conflicting optimizations among weather scenarios, leading to unstable performance. Therefore, addressing weather-specific discrepancies is essential for maintaining robust and consistent detection across all conditions.

\subsection{Mixture of Experts (MoE).}
MoE~\cite{Moe-llava, moe_survey, routing} has emerged as a powerful framework for scaling models while maintaining computational efficiency. Initially proposed by~\cite{adaptive_moe}, MoE divides the model into specialized experts and uses a gating network to select the most relevant experts for each input. Sparsely-Gated MoE~\cite{sparse_moe} further improves scalability by activating only a subset of experts, allowing models to scale to billions of parameters without significant computational overhead. GShard~\cite{Gshard}optimized MoE training on distributed systems, enabling efficient large-scale training. Switch Transformer~\cite{Switch_transformers} simplified expert routing by adopting top-1 selection, enhancing both training stability and scalability. Later works, such as GLaM~\cite{GLaM} and DeepSpeed-MoE~\cite{deepspeed_moe}, focused on improving MoE for multi-task learning and large-scale training. In contrast, V-MoE~\cite{V_MoE} extended MoE to vision tasks by applying sparse activation to image patches in Vision Transformers~\cite{ViT}, thereby improving computational efficiency.

The MoE framework offers a promising solution to the challenges posed by diverse data distributions in tasks with varying conditions. Motivated by these advantages, we are the first to introduce MoE into 3D object detection under adverse weather conditions, effectively addressing inter-weather discrepancies and enabling robust performance across all conditions.

\section{Proposed method}
\label{sec:method}

\begin{figure*}[ht]
  \centering
   \includegraphics[width=\linewidth]{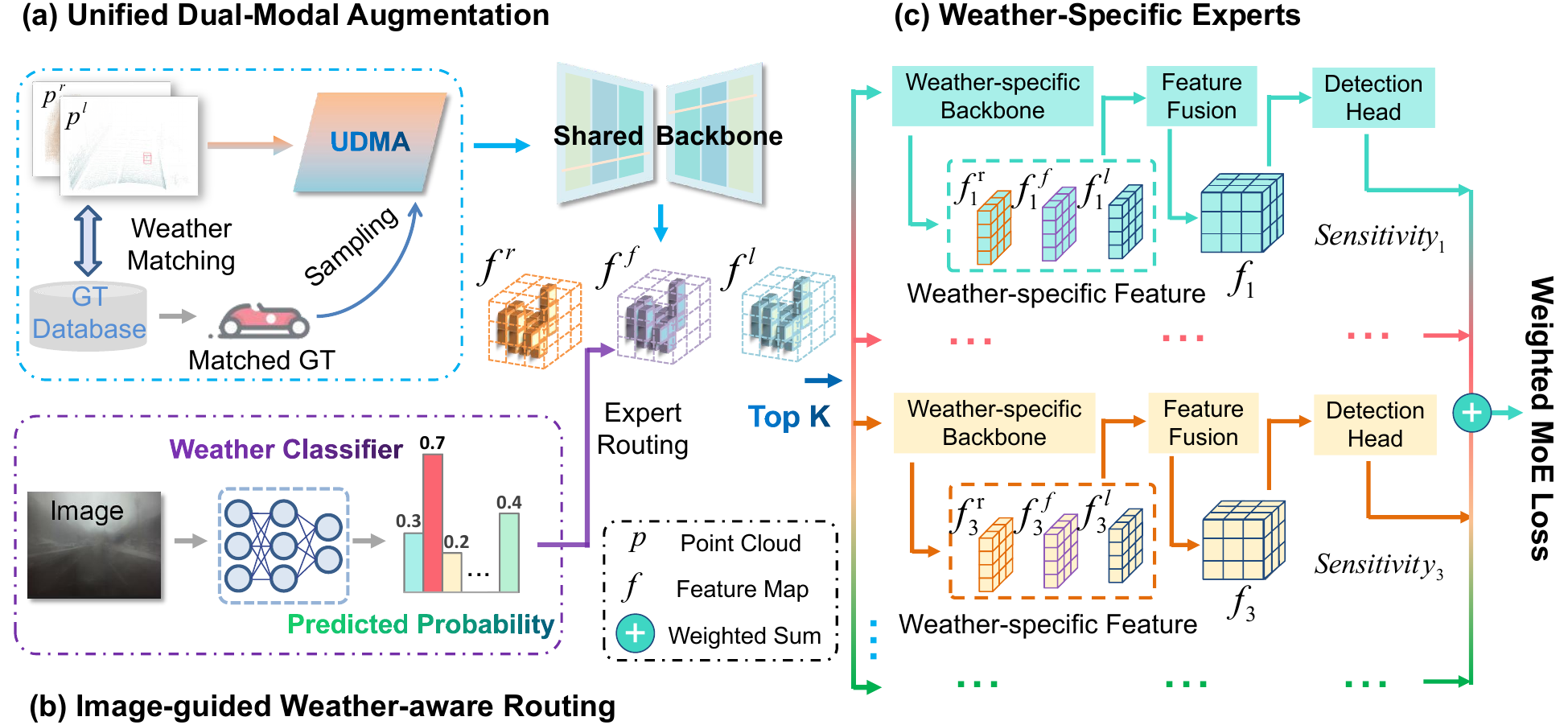}
   \caption{AW-MoE Framework. (a) Unified Dual-Modal Augmentation (UDMA): Synchronously augments LiDAR and 4D Radar point clouds. Its GT Sampling only selects ground truths matching the scene's weather. (b) Image-guided Weather-aware Routing (IWR): Uses an Image-based Weather Classifier to predict the scene weather and routes the feature to the top-K most relevant Weather-Specific Experts. (c) Weather-Specific Experts (WSE): Each expert is specialized for a weather condition, extracting robust weather-specific features and regressing bounding boxes with tailored sensitivity.}
   \label{fig:MoE}
\end{figure*}

\subsection{Problem Formulation}
In outdoor adverse weather scenarios, the sensory inputs, denoted as $\mathcal{I}$, are processed by a perception model $\mathcal{M}$ to extract deep representations $f = \mathcal{M}(\mathcal{I})$. For multi-modal settings~\cite{l4dr, towards}, features $f$ from different sensors are further integrated by a feature fusion module $\mathcal{G}$, producing the fused representation $f' = \mathcal{G}(\{f\})$.
The fused features are then fed into the detection head to regress the final 3D bounding boxes $B = \{b_i\}^{N_{b}}_{i=1},B \in\mathbb{R}^{N_b \times 7}$, where $N_b$ denotes the number of detected bounding boxes.

Our proposed AW-MoE builds upon the state-of-the-art LiDAR–4D Radar fusion framework L4DR~\cite{l4dr} by integrating a Mixture of Experts (MoE) mechanism. The input consists of LiDAR point clouds $\mathcal{P}^l$ and 4D Radar point clouds $\mathcal{P}^r$, denoted collectively as $\mathcal{P}^m = \{p_i^m\}_{i=1}^{N_m}, \ m \in \{l, r\}$, where $p_i^m$ represents a 3D point in modality $m$.

\subsection{AW-MoE}
The overall architecture of AW-MoE is illustrated in Fig.~\ref{fig:MoE}. AW-MoE consists of three main components: a Shared Backbone, an Image-guided Weather-aware Routing (IWR) module, and multiple Weather-Specific Experts (WSE). The Shared Backbone extracts general representations from the input data, while the IWR leverages discriminative visual cues from images under different weather conditions to dynamically route features to the most suitable WSE. Each WSE is specialized in processing features corresponding to a particular weather type, enabling AW-MoE to maintain robust and consistent detection performance across diverse adverse conditions. Moreover, the proposed Unified Dual-Modal Augmentation (UDMA) performs synchronized data augmentation for both LiDAR and 4D Radar modalities, ensuring sample authenticity and cross-modal consistency under various weather scenarios.

\subsubsection{Unified Dual-Modal Augmentation} 
Data augmentation~\cite{pointpillars, centerpoint, voxel_RCNN} is widely used in deep learning but has been largely overlooked in LiDAR–4D Radar fusion~\cite{l4dr, all-weather, towards}. In this work, we address this limitation by proposing Unified Dual-Modal Augmentation (UDMA), which performs synchronized augmentations on LiDAR and 4D Radar data, including flipping, rotation, scaling, and ground-truth (GT) sampling, to maintain cross-modal consistency. Unlike conventional GT sampling~\cite{pv-rcnn, second} which indiscriminately mixes data from different weather conditions and thereby degrades scene realism, our proposed Weather-Specific GT Sampling (WSGTS) accounts for the substantial geometric and reflective variations of objects across diverse weather scenarios. As shown in Fig.~\ref{fig:MoE} (a), WSGTS samples GTs exclusively from scenes with matching weather conditions, effectively avoiding cross-weather mismatches, preserving environmental authenticity, and improving detection performance, as reported in Table~\ref{tab:GT Sampling}.

\subsubsection{Image-guided Weather-aware Routing} 
The key to the MoE framework’s effectiveness in handling multi-task and multi-scenario problems lies in the ability of expert routing to accurately select the most suitable expert. As analyzed in the Introduction, the Point-cloud Feature-based Routing (PFR)~\cite{adaptive_moe}, which relies on point cloud features, performs poorly in outdoor scenarios due to the highly dynamic nature of environments and the difficulty of capturing point cloud differences under adverse weather such as fog, sleet, and light snow (see Figs.~\ref{fig:Image Advantage} (c, d) and Table~\ref{tab:Routing}).

Conversely, images offer superior clarity in distinguishing diverse weather patterns while remaining largely invariant to fluctuations in scene geometry (see Figs.~\ref{fig:Image Advantage} (a, b)). Based on this observation, we design an Image-guided Weather-aware Routing (IWR) to perform expert selection. First, we design a lightweight image-based Weather Classifier to categorize the captured scene images:
\begin{equation}
  z = \mathcal{C}(\mathcal{I}_{img}) \in \mathbb{R}^{N_{W}},
  \label{eq:classifier logits}
\end{equation}
where $z$ denotes the classification result, $\mathcal{C}$ represents the Weather Classifier, $\mathcal{I}_{img}$ denotes camera image, and $N_W$ is the number of weather categories. Then, the classification result $z$ is normalized using a softmax function, where $P_w$ denotes the probability corresponding to the $w$-th weather category:
\begin{equation}
P = \mathrm{softmax}(z), \\
P_w = \frac{\mathrm{exp}(z_w)}{\sum_{i=1}^{N_W}\mathrm{exp}(z_i)},\ w=1,...,N_W.
  \label{eq:softmax}
\end{equation}
Finally, we select the top-$K$ weather categories with the highest probabilities in $P$ to determine the corresponding Weather-Specific Experts (WSE):
\begin{equation}
  \mathcal{S} = \mathrm{TopK}(P,K) \subset {1,...,N_W}, \ \ \ \left \vert \mathcal{S} \right \vert = K,
  \label{eq:Top K}
\end{equation}
where $\mathcal{S}$ denotes the set of selected WSE. Since the proposed lightweight image-based Weather Classifier achieves high accuracy in predicting scene weather types (over 99\%, see Table~\ref{tab:Routing}), our IWR can reliably select the most appropriate WSE.

\begin{figure}[t]
  \centering
   \includegraphics[width=0.85\linewidth]{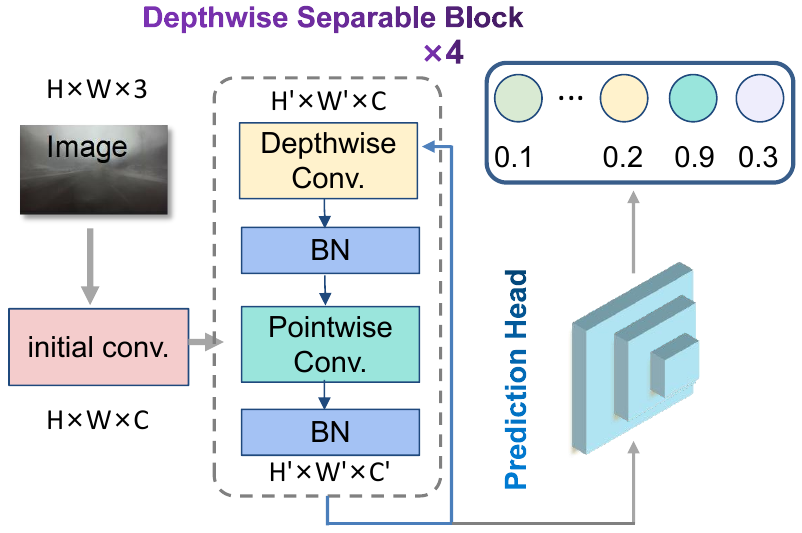}
   \caption{Architecture of the proposed Image-based Weather Classifier.}
   \label{fig:Weather Classifier}
\end{figure}

\textbf{Weather Classifier.}
The architecture of our Image-based Weather Classifier is illustrated in Fig.~\ref{fig:Weather Classifier}. It consists of an initial convolutional layer followed by a backbone composed of four consecutive Depthwise Separable Blocks. Each Depthwise Separable Block contains a depthwise convolution, a pointwise convolution, and two normalization layers, which collaboratively extract discriminative weather-related features from the input image. Despite its lightweight design, the proposed image-based Weather Classifier achieves both high efficiency and accuracy, providing precise and efficient routing for weather-specific experts.

\begin{figure*}[ht]
  \centering
   \includegraphics[width=\linewidth]{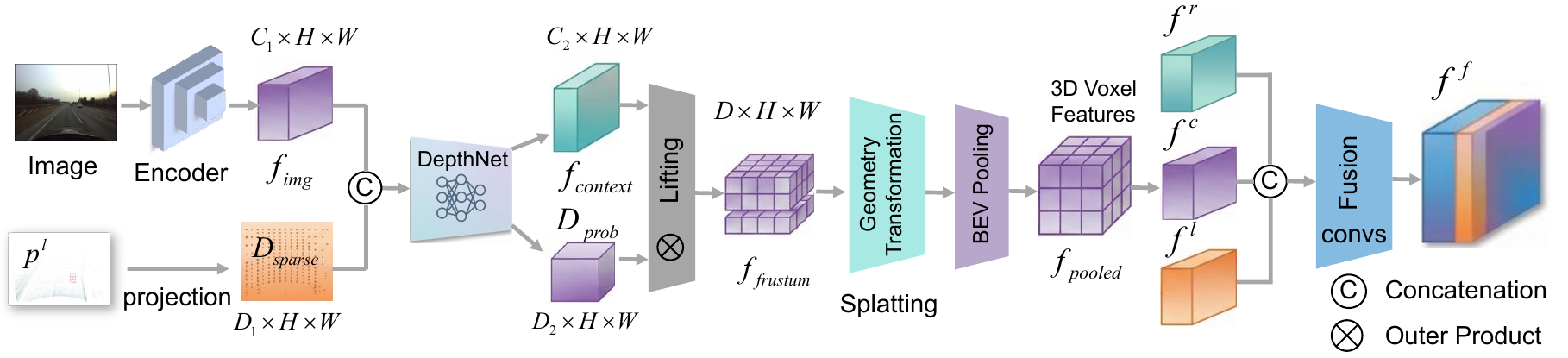}
   \caption{The architecture of the AW-MoE-LRC framework. The pipeline comprises three stages: (i) LiDAR-Guided Image Feature Lifting, where sparse LiDAR depth assists in predicting 3D frustum features from images; (ii) 3D Geometry Transformation and BEV Pooling, which projects and aggregates these features into the ego-vehicle BEV space; and (iii) Multi-Modal Feature Fusion, which concatenates the aligned camera, LiDAR, and 4D Radar BEV features along the channel dimension for final convolution-based integration.}
   \label{fig:Image Fusion}
\end{figure*}

\subsubsection{Weather-Specific Experts} 
After IWR selects the most suitable expert, the corresponding Weather-Specific Expert (WSE) is activated to handle the scenario under a specific weather condition. Each WSE consists of three components: a Weather-Specific Backbone, a Weather-Specific Feature Fusion module, and a Weather-Specific Detection Head. The Weather-Specific Backbone is responsible for extracting weather-specific features that cannot be captured by the shared backbone. The Weather-Specific Feature Fusion module performs weather-aware complementary fusion of LiDAR and 4D Radar features according to their quality differences under different weather conditions. The Weather-Specific Detection Head predicts and regresses 3D bounding boxes with varying sensitivities tailored to specific weather scenarios.
The overall pipeline of WSE can be formulated as:
\begin{equation}
B_w = \mathcal{H}_w(\mathcal{F}_w(\mathcal{E}_w(\{f\}))), \quad w \in \mathcal{S},
\label{eq:WSE}
\end{equation}
where $\mathcal{E}_w$, $\mathcal{F}_w$, and $\mathcal{H}_w$ denote the $w$-th Weather-Specific Backbone, Weather-Specific Feature Fusion module, and Weather-Specific Detection Head, respectively. In AW-MoE, all WSEs share the same structural design, with a total of $N_W = 7$ experts corresponding to the number of weather categories.

\begin{table}[t]
\caption{Comparison of weather classification accuracy between Point-cloud Feature-based Routing (PFR) and the proposed Image-guided Weather-aware Routing (IWR).}
\centering
\setlength{\tabcolsep}{1pt}
\resizebox{\columnwidth}{!}{

\begin{tabular}{l|cccccccc}
\hline
Method            & Total         & Normal        & Overcast       & Fog           & Rain          & Sleet         & Light Snow     & Heavy Snow     \\ \hline
PFR    & 71.3          & 98.9          & 12.0           & 51.9          & 76.7          & 57.6          & 2.1           & 58.3          \\
IWR & \textbf{99.0} & \textbf{99.8} & \textbf{100.0} & \textbf{97.3} & \textbf{99.0} & \textbf{97.1} & \textbf{99.3} & \textbf{98.7} \\ \hline
\end{tabular}

}
  \label{tab:Routing}%
\end{table}

\subsection{AW-MoE-LRC: Integrating Image Features}
In the AW-MoE framework (Fig.~\ref{fig:MoE}), camera images are exclusively used within the Image-guided Weather-aware Routing (IWR) module to select Weather-Specific Experts. The image features are not directly utilized for 3D object detection. To explicitly integrate image semantics with LiDAR and 4D Radar features, we propose an extended pipeline, AW-MoE-LRC, as illustrated in Fig.~\ref{fig:Image Fusion}. Following \cite{bevfusion}, we adopt a LiDAR-guided Lift-Splat-Shoot (LSS) architecture to map 2D image features into a unified Bird's-Eye-View (BEV) space for multi-modal alignment. This process consists of three main stages:

\subsubsection{LiDAR-Guided Image Feature Lifting}
Standard LSS architectures often lack precise geometric constraints for depth estimation. To address this, we leverage sparse LiDAR point clouds to guide image depth prediction. First, we project the LiDAR point clouds $\mathcal{P}^l$ onto the camera image plane using the intrinsic matrix $A$ and extrinsic matrix $T_{ext}$ to generate a sparse depth map $D_{sparse}  \in \mathbb{R}^{D_1 \times H \times W}$. This depth map is convolved, concatenated with the backbone-extracted image features $f_{img}$, and fed into a DepthNet. The network outputs context features $f_{context} \in \mathbb{R}^{C_2 \times H \times W}$ and a discrete depth probability distribution $D_{prob} \in \mathbb{R}^{D_2 \times H \times W}$, where $D_2$ denotes the number of predefined depth bins. The 3D frustum feature $f_{frustum}$ is then computed via the outer product of the depth probabilities and context features:
\begin{equation}
  f_{frustum}(u, v, d) = D_{prob}(u, v, d) \otimes f_{context}(u, v),
  \label{eq:lifting}
\end{equation}
where $(u, v)$ represents the image pixel coordinates and $d$ is the discrete depth index.

\subsubsection{3D Geometry Transformation and BEV Pooling (Splatting)}
To map the frustum features into the ego-vehicle coordinate system, we compute the 3D coordinate $P_{ego}$ for each feature point. Given the depth $d$ and pixel coordinate $(u, v)$, and accounting for data augmentations (e.g., image augmentation matrix $T_{img\_aug}$ and LiDAR augmentation matrix $T_{lidar\_aug}$), the coordinate transformation is formulated as:
\begin{equation}
  P_{ego} = T_{lidar\_aug} \left( T_{ext} \ A^{-1} \ T_{img\_aug}^{-1} \begin{bmatrix} u \cdot d \\ v \cdot d \\ d \\ 1 \end{bmatrix} \right).
  \label{eq:splatting}
\end{equation}
After obtaining the 3D coordinates for all frustum features, we apply an efficient BEV pooling operation to aggregate features that fall into the same 3D voxel grid. The features along the $Z$-axis are then flattened and concatenated across the channel dimension. Finally, a downsampling convolutional layer is applied to generate the spatial BEV features for the camera branch, denoted as $f^{c}$.

\subsubsection{Multi-Modal Feature Fusion}
Once the image spatial features $f^{c}$ are extracted, they are fused with the LiDAR features $f^{l}$ and 4D Radar features $f^{r}$ within the unified BEV space. We concatenate the features along the channel dimension and apply several convolutional layers to learn cross-modal interactions and adaptive weight assignments. The final fused feature $f^{f}$ is obtained as follows:
\begin{equation}
  f^f = \text{Convs} \left( [f^{c}, f^{l}, f^{r}] \right), 
  \label{eq:image_fusion}
\end{equation}
where $[\cdot]$ denotes the channel-wise concatenation. This fusion strategy effectively harnesses the rich semantic information from images, the precise geometric structure of LiDAR, and the robust, all-weather dynamic perception of 4D Radar.

\begin{algorithm}[t]
\caption{AW-MoE Training Strategy}
\label{alg:awmoe_training}
\KwIn{LiDAR point clouds $\mathcal{P}^l$, 4D Radar point clouds $\mathcal{P}^r$, camera images $\mathcal{I}_{img}$}
\KwOut{Trained AW-MoE model}

\textbf{Stage 1: Pretrain single-branch AW-MoE}\;
Select a designated WSE$_d$\;
\For{each batch in all-weather data $\{\mathcal{P}^l, \mathcal{P}^r\}$}{
    Forward: $B_d \gets \mathcal{H}_d(\mathcal{F}_d(\mathcal{E}_d(\mathcal{E}_{shared}(\mathcal{P}^l, \mathcal{P}^r))))$\;
    Compute loss and update parameters of $\mathcal{E}_{shared}$ and WSE$_d$\;
}

\textbf{Stage 2: Train image-based Weather Classifier $\mathcal{C}$}\;
\For{each batch of $\mathcal{I}_{img}$ with weather labels}{
    Forward: $z \gets \mathcal{C}(\mathcal{I}_{img}) \in \mathbb{R}^{N_W}$\;
    Compute classification loss and update $\mathcal{C}$\;
}

\textbf{Stage 3: Initialize AW-MoE}\;
Freeze parameters of $\mathcal{E}_{shared}$\;
Copy pretrained parameters to all WSE branches:
$\text{WSE}_w \leftarrow \text{WSE}_d,\quad w = 1, \ldots, N_W$\;

\textbf{Stage 4: Train AW-MoE with IWR}\;
\For{each batch $\{\mathcal{P}^l, \mathcal{P}^r, \mathcal{I}_{img}\}$}{
    Extract shared features: $f \gets \mathcal{E}_{shared}(\mathcal{P}^l, \mathcal{P}^r)$\;
    Compute weather probabilities: $P \gets \mathrm{softmax}(\mathcal{C}(\mathcal{I}_{img})) \in \mathbb{R}^{N_W}$\;
    Select top-$K$ experts: $\mathcal{S} \gets \mathrm{TopK}(P, K)$\;
    Predict 3D boxes: $B_w \gets \mathcal{H}_w(\mathcal{F}_w(\mathcal{E}_w(\{f\}))), \quad w \in \mathcal{S}$\;
    Compute confidence-weighted loss: $\mathcal{L}_\mathcal{CW} \gets \sum_{w \in \mathcal{S}} P_w \, \mathcal{L}_w(\text{WSE}_w)$\;
    Update parameters of selected experts $\text{WSE}_w, \ w \in \mathcal{S}$\;
}
\end{algorithm}

\subsection{Loss Function and Post-Processing}
In the AW-MoE framework, the IWR selects the top-$K$ Weather-Specific Experts (WSE) to process the input data. During training, each selected WSE computes an individual loss, while during inference, each WSE regresses a dedicated set of 3D bounding boxes. However, the relevance between a WSE and the input data fluctuates based on weather conditions. To account for this varying contribution, a specialized loss function and post-processing strategy are required to aggregate the outputs. We thus propose the following formulations:

\subsubsection{Confidence-Weighted MoE Loss}
To account for the varying relevance between data and experts, we introduce the Confidence-Weighted MoE Loss. This objective function leverages the routing probabilities $P$, generated by the IWR, as dynamic confidence scores. The total loss is formulated as a weighted sum over the set of selected experts $S$:
\begin{equation}
  \mathcal{L}_\mathcal{CW} = \sum_{w \in \mathcal{S}} P_w \, \mathcal{L}_w(WSE_w),
  \label{eq:Loss}
\end{equation}
where $\mathcal{L}_w$ denotes the individual loss computed by the $w$-th WSE. Scaling each expert's contribution proportional to its routing probability $P_w$ prevents samples with low relevance from disproportionately affecting the optimization of specialized experts, thereby ensuring stable, weather-aware convergence.

\subsubsection{Confidence-Weighted Post-Processing} Complementing the weighted loss, we apply a consistent Confidence-Weighted Post-Processing during inference to aggregate the 3D bounding boxes $B = \{b_i\}_{i=1}^{N_b}$ regressed by the top-$K$ experts. This process effectively integrates multi-expert predictions through two stages: Candidate Selection and Confidence-Weighted Aggregation.

\textbf{Candidate Selection.} We first evaluate the 3D Intersection over Union (IoU) among all predicted boxes. Candidates with an IoU below a predefined matching threshold are retained as independent detections. \textbf{Confidence-Weighted Aggregation.} For overlapping boxes representing the same target, we perform a weighted aggregation. Let $\Omega$ denote a set of matched boxes, where each box $b_j \in \Omega$ is associated with its corresponding routing probability $p_j$. The fused bounding box $\hat{b}$ is 
\begin{equation}
  \hat{b} = \sum_{b_j \in \Omega} p_j \cdot b_j, \quad b_j \in \mathbb{R}^7
  \label{eq:post-processing}
\end{equation}
By sharing the same IWR-derived weights as the loss function, this post-processing module dynamically prioritizes predictions from experts most relevant to the current weather, ensuring robust and spatially consistent final detections.

\begin{table*}[ht]
\caption{Quantitative results of different 3D object detection methods on K-Radar dataset. We present the modality of each method (L: LiDAR, 4DR: 4D Radar) and detailed performance for each weather condition. Best in \textbf{bold}, second in \underline{underline}, and $^*$ indicates results reproduced using open code.}
\centering
\setlength{\tabcolsep}{6.5pt}
\resizebox{\textwidth}{!}{

\begin{tabular}{cc|cccccccccc}
\hline
Method                                                                          & Modality               & IoU                  & Metric & Total         & Normal        & Overcast      & Fog           & Rain          & Sleet         & Light Snow     & Heavy Snow     \\ \hline
\multirow{4}{*}{\begin{tabular}[c]{@{}c@{}}RTNH~\cite{k-radar}\\ (NeurIPS 2022)\end{tabular}}  & \multirow{4}{*}{4DR}   & \multirow{2}{*}{0.3} & $AP_{BEV}$  & 41.1          & 41.0          & 44.6          & 45.4          & 32.9          & 50.6          & 81.5          & 56.3          \\
        &                        &                      & $AP_{3D}$   & 37.4          & 37.6          & 42.0          & 41.2          & 29.2          & 49.1          & 63.9          & 43.1          \\ 
        &                        & \multirow{2}{*}{0.5} & $AP_{BEV}$  & 36.0          & 35.8          & 41.9          & 44.8          & 30.2          & 34.5          & 63.9          & 55.1          \\
        &                        &                      & $AP_{3D}$   & 14.1          & 19.7          & 20.5          & 15.9          & 13.0          & 13.5          & 21.0          & 6.36          \\ \hline
\multirow{4}{*}{\begin{tabular}[c]{@{}c@{}}RTNH~\cite{k-radar} \\ (NeurIPS 2022)\end{tabular}} & \multirow{4}{*}{L}     & \multirow{2}{*}{0.3} & $AP_{BEV}$ & 76.5          & 76.5          & 88.2          & 86.3          & 77.3          & 55.3          & 81.1          & 59.5          \\
        &                        &                      & $AP_{3D}$   & 72.7          & 73.1          & 76.5          & 84.8          & 64.5          & 53.4          & 80.3          & 52.9          \\ 
        &                        & \multirow{2}{*}{0.5} & $AP_{BEV}$  & 66.3          & 65.4          & 87.4          & 83.8          & 73.7          & 48.8          & 78.5          & 48.1          \\
        &                        &                      & $AP_{3D}$   & 37.8          & 39.8          & 46.3          & 59.8          & 28.2          & 31.4          & 50.7          & 24.6          \\ \hline
\multirow{4}{*}{\begin{tabular}[c]{@{}c@{}}InterFusion$^*$~\cite{interfusion} \\ (IROS 2023)\end{tabular}}  & \multirow{4}{*}{L+4DR} & \multirow{2}{*}{0.3} & $AP_{BEV}$  & 69.5    & 76.6          & 84.9          & 84.3    & 70.2         & 35.1          & 63.1          & 46.3 \\
        &                        &                      & $AP_{3D}$   & 65.6          & 72.5
    & 81.4    & 76.9         & 63.8         & 34.6          & 59.9   & 45.9 \\ 
        &                        & \multirow{2}{*}{0.5} & $AP_{BEV}$  & 66.1          & 70.5         & 82.0         & 81.8         & 67.2         & 33.9         & 62.9         & 46.0    \\
        &                        &                      & $AP_{3D}$   & 41.7          & 44.6         & 53.5          & 64.8         & 37.2          & 25.5          & 35.4    & 27.0          \\ \hline
\multirow{4}{*}{\begin{tabular}[c]{@{}c@{}}3D-LRF~\cite{towards} \\ (CVPR 2024)\end{tabular}}  & \multirow{4}{*}{L+4DR} & \multirow{2}{*}{0.3} & $AP_{BEV}$  & {\ul 84.0}    & 83.7          & 89.2          & {\ul 95.4}    & 78.3          & 60.7          & 88.9          & \textbf{74.9} \\
        &                        &                      & $AP_{3D}$   & 74.8          & 81.2    & 87.2    & 86.1          & 73.8          & 49.5          & {\ul 87.9}    & \textbf{67.2} \\  
        &                        & \multirow{2}{*}{0.5} & $AP_{BEV}$  & 73.6          & 72.3          & 88.4          & 86.6          & 76.6          & 47.5          & 79.6          & {\ul 64.1}    \\
        &                        &                      & $AP_{3D}$   & 45.2          & 45.3          & 55.8          & 51.8          & 38.3          & 23.4          & {\ul 60.2}    & 36.9          \\ \hline
\multirow{4}{*}{\begin{tabular}[c]{@{}c@{}}L4DR~\cite{l4dr}\\ (AAAI 2025)\end{tabular}}     & \multirow{4}{*}{L+4DR} & \multirow{2}{*}{0.3} & $AP_{BEV}$  & 79.5          & 86.0    & 89.6    & 89.9          & {\ul 81.1}    & 62.3    & 89.1    & 61.3          \\
        &                        &                      & $AP_{3D}$   & 78.0    & 77.7          & 80.0          & 88.6    & 79.2    & 60.1    & 78.9          & 51.9          \\ 
        &                        & \multirow{2}{*}{0.5} & $AP_{BEV}$  & 77.5    & 76.8    & 88.6    & 89.7    & 78.2    & {\ul 59.3}    & 80.9    & 53.8          \\
        &                        &                      & $AP_{3D}$   & 53.5    & 53.0    & 64.1    & 73.2    & 53.8    & {\ul 46.2} & 52.4          & 37.0    \\ \hline
\multirow{4}{*}{\begin{tabular}[c]{@{}c@{}}L4DR-DA3D~\cite{DA3D}\\ (MM 2025)\end{tabular}}     & \multirow{4}{*}{L+4DR} & \multirow{2}{*}{0.3} & $AP_{BEV}$  & 80.4          & {\ul 86.5}    & {\ul 89.8}    & 90.1          & 81.0    & {\ul 62.6}    & {\ul 89.9}    & 61.9          \\
        &                        &                      & $AP_{3D}$   & {\ul 79.3}    & \textbf{85.9}          & {\ul 88.4}          & {\ul 89.2}    & {\ul 79.7}    & {\ul 65.8}    & {\ul 89.0}         & 60.2          \\ 
        &                        & \multirow{2}{*}{0.5} & $AP_{BEV}$  & {\ul 78.5}    & {\ul 77.4}    & {\ul 89.1}    & {\ul 90.1}    & {\ul 79.3}    & 58.8    & {\ul 88.9}    & 60.6          \\
        &                        &                      & $AP_{3D}$   & \textbf{61.9}    & {\ul 58.9}    & {\ul 66.4}    & {\ul 79.2}    & {\ul 63.0}    & \textbf{48.2} & {\ul 64.6}         & {\ul 47.6}    \\ \hline
\multirow{4}{*}{\begin{tabular}[c]{@{}c@{}}AW-MoE\\ (Ours)\end{tabular}}                                                           & \multirow{4}{*}{L+4DR} & \multirow{2}{*}{0.3} & $AP_{BEV}$  & \textbf{88.2} & \textbf{87.7} & \textbf{94.5} & \textbf{96.7} & \textbf{88.8} & \textbf{81.0} & \textbf{95.4} & {\ul 70.2}    \\
        &                        &                      & $AP_{3D}$   & \textbf{83.9} & {\ul 84.2} & \textbf{90.0} & \textbf{95.3} & \textbf{84.4} & \textbf{72.9} & \textbf{90.2} & {\ul 64.0}          \\ 
        &                        & \multirow{2}{*}{0.5} & $AP_{BEV}$  & \textbf{84.2} & \textbf{82.8} & \textbf{91.6} & \textbf{96.3} & \textbf{85.3} & \textbf{75.0} & \textbf{94.7} & \textbf{66.4} \\
        &                        &                      & $AP_{3D}$   & {\ul 61.5} & \textbf{59.0} & \textbf{67.2} & \textbf{85.7} & \textbf{63.5} & 43.3    & \textbf{70.1} & \textbf{53.1} \\ \hline
\end{tabular}

}
  \label{tab:MoE_Table}%
\end{table*}

\begin{table*}[ht]
\caption{Performance ($AP_{3D}$) of AW-MoE and its camera-integrated variant, AW-MoE-LRC.}
\centering
\setlength{\tabcolsep}{7pt}
\resizebox{\textwidth}{!}{

\begin{tabular}{ll|cccccccccc}
\hline
Method                      & Modality                 & IoU & Total & Normal & Overcast & Fog  & Rain & Sleet & Light Snow & Heavy Snow & FPS {[}HZ{]}            \\ \hline
\multirow{2}{*}{AW-MoE}     & \multirow{2}{*}{L+4DR}   & 0.3 & 83.9  & 84.2   & 90.0     & 95.3 & 84.4 & 72.9  & 90.2       & 64.0       & \multirow{2}{*}{\textbf{12.41}} \\
    &                          & 0.5 & 61.5  & 59.0   & 67.2     & 85.7 & 63.5 & 43.3  & 70.1       & 53.1       &                        \\ \hline
\multirow{2}{*}{AW-MoE-LRC} & \multirow{2}{*}{L+4DR+C} & 0.3 & 84.3  & \textbf{84.7}   & \textbf{91.0}     & 95.3 & 84.0 & 72.9  & 89.6       & 63.7       & \multirow{2}{*}{10.02}  \\
    &                          & 0.5 & 61.8  & \textbf{60.2}   & \textbf{70.4}     & 85.8 & 63.4 & 43.7  & 69.9       & 52.8       &                        \\ \hline
\end{tabular}

}
  \label{tab:Image-fusion}%
\end{table*}

\begin{table*}[ht]
\caption{Performance ($AP_{3D}$) comparison of AW-MoE when extended to different 3D object detection baselines.}
\centering
\setlength{\tabcolsep}{8pt}
\resizebox{\textwidth}{!}{

\begin{tabular}{lccccccccc}
\hline
Method                                                      & IoU & Total & Normal & Overcast & Fog  & Rain & Sleet & Light Snow & Heavy Snow \\ \hline
\multicolumn{1}{l|}{\multirow{2}{*}{RTNH (4DR)~\cite{k-radar}}}            & 0.3 & 37.4  & 37.6   & 42.0     & 41.2 & 29.2 & 49.1  & 63.9      & 43.1      \\
\multicolumn{1}{l|}{}                                       & 0.5 & 14.1  & 19.7   & 20.5     & 15.9 & 13.0 & 13.5  & 21.0      & 6.4       \\
\multicolumn{1}{l|}{\multirow{2}{*}{RTNH (4DR) - AW-MoE}}   & 0.3 & 65.7  & 64.4   & 72.2     & 88.7 & 58.3 & 64.0  & 71.2      & 65.0      \\
\multicolumn{1}{l|}{}                                       & 0.5 & 35.7  & 28.6   & 44.3     & 72.9 & 34.9 & 32.4  & 42.7      & 45.3      \\ 
\rowcolor[gray]{0.9} \multicolumn{1}{l|}{}           & \textit{0.3} & \textit{+28.3}  & \textit{+26.8}   & \textit{+30.2}     & \textit{+47.5} & \textit{+29.1} & \textit{+14.9}  & \textit{+7.3}       & \textit{+21.9}      \\
\rowcolor[gray]{0.9} \multicolumn{1}{l|}{\multirow{-2}{*}{\textit{Improvement}}}           & \textit{0.5} & \textit{+21.6}  & \textit{+8.9}    & \textit{+23.8}     & \textit{+57.0} & \textit{+21.9} & \textit{+18.9}  & \textit{+21.7}      & \textit{+38.9}      \\ \hline
\multicolumn{1}{l|}{\multirow{2}{*}{RTNH (L)~\cite{k-radar}}}          & 0.3 & 72.7  & 73.1   & 76.5     & 84.8 & 64.5 & 53.4  & 80.3      & 52.9      \\
\multicolumn{1}{l|}{}                                       & 0.5 & 37.8  & 39.8   & 46.3     & 59.8 & 28.2 & 31.4  & 50.7      & 24.6      \\
\multicolumn{1}{l|}{\multirow{2}{*}{RTNH (L) - AW-MoE}} & 0.3 & 81.1  & 84.0   & 89.6     & 93.5 & 83.4 & 56.2  & 89.4      & 57.3      \\
\multicolumn{1}{l|}{}                                       & 0.5 & 55.4  & 53.0   & 51.2     & 82.1 & 59.4 & 38.7  & 69.5      & 40.5      \\ 
\rowcolor[gray]{0.9} \multicolumn{1}{l|}{}           & \textit{0.3} & \textit{+8.4}   & \textit{+10.9}   & \textit{+13.1}     & \textit{+8.7}  & \textit{+18.9} & \textit{+2.8}   & \textit{+9.1}       & \textit{+4.4}       \\
\rowcolor[gray]{0.9} \multicolumn{1}{l|}{\multirow{-2}{*}{\textit{Improvement}}}       & \textit{0.5} & \textit{+17.6}  & \textit{+13.2}   & \textit{+4.9}      & \textit{+22.3} & \textit{+31.2} & \textit{+7.3}   & \textit{+18.8}      & \textit{+15.9}      \\ \hline
\multicolumn{1}{l|}{\multirow{2}{*}{InterFusion~\cite{interfusion}}}           & 0.3 & 65.6  & 72.5   & 81.4     & 76.9 & 63.8 & 34.6  & 59.9      & 45.9      \\
\multicolumn{1}{l|}{}                                       & 0.5 & 41.7  & 44.6   & 53.5     & 64.8 & 37.2 & 25.5  & 35.4      & 27.0      \\
\multicolumn{1}{l|}{\multirow{2}{*}{InterFusion - AW-MoE}}  & 0.3 & 81.7  & 83.8   & 89.7     & 92.7 & 80.4 & 71.2  & 82.5      & 60.7      \\
\multicolumn{1}{l|}{}                                       & 0.5 & 60.3  & 58.0   & 70.1     & 85.9 & 60.9 & 47.0  & 63.4      & 44.7      \\ 
\rowcolor[gray]{0.9} \multicolumn{1}{l|}{}           & \textit{0.3} & \textit{+16.1}  & \textit{+11.3}   & \textit{+8.3}      & \textit{+15.8} & \textit{+16.6} & \textit{+36.6}  & \textit{+22.6}      & \textit{+14.8}      \\
\rowcolor[gray]{0.9} \multicolumn{1}{l|}{\multirow{-2}{*}{\textit{Improvement}}}        & \textit{0.5} & \textit{+18.6}  & \textit{+13.4}   & \textit{+16.6}     & \textit{+21.1} & \textit{+23.7} & \textit{+21.5}  & \textit{+28.0}      & \textit{+17.7}      \\ \hline
\end{tabular}

}
  \label{tab:Generalization}%
\end{table*}

\begin{table}[t]
\caption{FPS, and FLOPS comparison of detectors before and after applying AW-MoE (corresponding to Table~\ref{tab:Generalization}).}
\centering
\setlength{\tabcolsep}{1.5pt}
\resizebox{\columnwidth}{!}{

\begin{tabular}{l|c|c|c|c}
\hline
Method               & Sensors                & FPS {[}HZ{]} & FLOPS {[}GB{]} & Param {[}M{]} \\ \hline
L4DR~\cite{l4dr}                 & \multirow{2}{*}{L+4DR} & 13.94        & 142.65         & 59.73            \\
L4DR - AW-MoE        &                        & 12.41        & 143.41         & 143.46           \\ \hline
RTNH (4DR)~\cite{k-radar}           & \multirow{2}{*}{4DR}   & 14.77        & 502.57         &   17.35            \\
RTNH (4DR)  - AW-MoE &                        & 14.50        & 503.25         &  61.41             \\ \hline
RTNH (L)~\cite{k-radar}             & \multirow{2}{*}{L}     & 14.62        & 502.53         &     17.35          \\
RTNH (L)  - AW-MoE   &                        & 14.20        & 503.22         &        61.41
       \\ \hline
InterFusion~\cite{interfusion}          & \multirow{2}{*}{L+4DR} & 31.64        & 16.72          &       3.87        \\
InterFusion - AW-MoE &                        & 29.94        & 17.48          &      20.71         \\ \hline
\end{tabular}

}
  \label{tab:FPS FLOPS}%
\end{table}

\subsection{Training Strategy}
As mentioned in the Introduction, collecting data under adverse weather conditions is challenging, resulting in significantly fewer samples for each adverse condition (see Fig.~\ref{fig:Data} (b)). Even with top-$K$ expert routing, some Weather-Specific Experts may not receive sufficient training. To address this, we propose a training strategy tailored for AW-MoE, as summarized in Algorithm~\ref{alg:awmoe_training}. First, all weather data are used to train a single WSE branch, allowing the model to acquire basic 3D object detection capabilities. Next, the Shared Backbone is frozen, and the trained parameters of this WSE are copied to each branch for further training. Combined with the top-$K$ expert routing, this strategy effectively mitigates the training challenges caused by limited adverse-weather data.
\section{Experiments}
\label{sec:Experiments}

\subsection{Dataset and Evaluation Metrics}
The K-Radar dataset~\cite{k-radar} contains 58 sequences with a total of 34,944 frames (17,486 for training and 17,458 for testing), collected with 64-line LiDAR, cameras, and 4D Radar sensors. It includes not only normal conditions but also six types of adverse weather, such as fog, rain, and heavy snow. For evaluation, we adopt two standard metrics for 3D object detection: 3D Average Precision ($AP_{3D}$) and Bird’s Eye View Average Precision ($AP_{BEV}$), which are measured on the “Sedan” class at IoU thresholds of 0.3 and 0.5.

\subsection{Implement Details}
Our AW-MoE is designed as a general framework that can be extended to various 3D object detection algorithms. In this work, we extend the L4DR~\cite{l4dr} baseline to develop AW-MoE. Furthermore, we propose AW-MoE-LRC, which integrates camera image features into the AW-MoE framework. To achieve a balance between detection performance and inference efficiency, we set $K=1$ in the Image-guided Weather-aware Routing. The model is trained on four RTX 3090 GPUs with a batch size of 3.


\subsection{Results on K-Radar Adverse Weather Dataset}
Following L4DR~\cite{l4dr}, we compare AW-MoE with several modality-based 3D object detection methods, including RTNH~\cite{k-radar}, InterFusion~\cite{interfusion}, 3D-LRF~\cite{towards}, L4DR~\cite{l4dr} and L4DR-DA3D~\cite{DA3D}. The results are reported in Table~\ref{tab:MoE_Table}. AW-MoE consistently outperforms the state-of-the-art methods under both normal and adverse weather conditions, with particularly notable gains under extreme weather such as fog, sleet, light snow, and heavy snow. Specifically, compared to its baseline L4DR, our extended AW-MoE achieves a 10\% increase in $AP_{3D}$ (IoU=0.3) under fog, a 12.5\% increase in $AP_{3D}$ (IoU=0.5) under rain, a 12.8\% increase in $AP_{3D}$ (IoU=0.3) under sleet, and approximately 15\% improvements in $AP_{3D}$ (IoU=0.3 and 0.5) under light snow and heavy snow. Furthermore, our AW-MoE significantly outperforms L4DR-DA3D across most evaluation metrics. These improvements are attributed to AW-MoE’s multi-branch Weather-Specific Expert design, which mitigates performance conflicts arising from large inter-weather variations, and the precise expert selection enabled by the Image-guided Weather-aware Routing, which further enhances the model’s robustness across diverse conditions.

\subsection{Extensibility of AW-MoE to Other 3D Detectors}
To evaluate the extensibility of AW-MoE to other 3D object detectors, we applied it to RTNH~\cite{k-radar} and InterFusion~\cite{interfusion}, where RTNH includes both LiDAR and 4D Radar variants. As shown in Table~\ref{tab:Generalization}, incorporating AW-MoE consistently improves detection performance across various weather conditions and IoU thresholds, yielding improvements of over 15\%. Notably, after integrating AW-MoE, RTNH (4DR)~\cite{k-radar} and InterFusion~\cite{interfusion} outperform the state-of-the-art methods listed in Table~\ref{tab:MoE_Table}, enabling previously inferior models to surpass them; for instance, InterFusion achieves a 6.8\% higher total performance than L4DR in $AP_{3D}$ (IoU=0.5), with even larger gains under adverse weather. These results demonstrate that AW-MoE is highly compatible and effective across different detectors, further validating the robustness and generality of its design.

\subsection{Performance of AW-MOE-LRC}
Table~\ref{tab:Image-fusion} presents the evaluation results of our camera-integrated variant, AW-MoE-LRC, on the K-Radar dataset. Compared to AW-MoE, AW-MoE-LRC moderately improves detection accuracy under high-visibility conditions, such as normal and overcast weather. However, it yields negligible gains in severe weather like fog, rain, and snow. This occurs because camera sensors require clear visibility to capture useful semantic information; in extreme weather, degraded visibility renders these features ineffective for detection. These results underscore the strategic design of our IWR. By leveraging the distinct visual characteristics of images to classify weather and route inputs to the appropriate expert, IWR provides a much more effective way to utilize camera data.

\subsection{Computational Efficiency of AW-MoE}
The key advantage of the MoE framework lies in its multi-branch architecture, which effectively handles diverse tasks and scenarios. Since expert routing activates only a subset of experts during inference, it incurs only a minimal increase in computational cost. Our AW-MoE inherits this property. As shown in Table~\ref{tab:FPS FLOPS}, AW-MoE introduces negligible impact on inference speed and FLOPs when extended to different baselines. This efficiency stems from the lightweight design of the Image-guided Weather-aware Routing module, which precisely selects the appropriate experts while adding only marginal computational overhead. Furthermore, the table indicates that the parameter overhead introduced by this design remains within an acceptable range, ensuring its viability for practical deployment.

\begin{table*}[ht]
\caption{Performance comparison between Weather-Specific GT Sampling (WSGTS) and Weather-Agnostic GT Sampling (WAGTS). Non-normal denotes the aggregate of all non-normal weather conditions.}
\centering
\setlength{\tabcolsep}{7.5pt}
\resizebox{0.95\textwidth}{!}{

\begin{tabular}{cccccccccccc}
\hline
Method                                                             & IoU                  & Metric & Total         & Normal        & Overcast      & Fog           & Rain          & Sleet         & Light Snow     & Heavy Snow     & Non-normal      \\ \hline
\multicolumn{1}{c|}{\multirow{4}{*}{WAGTS}} & \multirow{2}{*}{0.3} & $AP_{BEV}$  & 87.4          & 87.2          & 94.7          & 96.3          & 87.9          & 78.8          & 94.7          & 71.7          & 86.9          \\
\multicolumn{1}{c|}{}                                              &                      & $AP_{3D}$   & 82.5          & 83.6          & 87.3          & 93.1          & 83.6          & 69.7          & 89.2          & 65.5          & 81.6          \\ 
\multicolumn{1}{c|}{}                                              & \multirow{2}{*}{0.5} & $AP_{BEV}$  & 83.2          & 82.1          & 90.3          & 96.1          & 84.4          & 70.4          & 93.8          & 67.6          & 83.4          \\
\multicolumn{1}{c|}{}                                              &                      & $AP_{3D}$   & 59.3          & 58.1          & 60.6          & 77.7          & 63.6          & 37.4          & 63.6          & 50.2          & 60.1          \\ \hline
\multicolumn{1}{c|}{\multirow{4}{*}{WSGTS}} & \multirow{2}{*}{0.3} & $AP_{BEV}$  & \textbf{88.2} & \textbf{87.7} & 94.4          & \textbf{96.5} & \textbf{88.3} & \textbf{79.5} & \textbf{95.4} & \textbf{72.8} & \textbf{87.9} \\
\multicolumn{1}{c|}{}                                              &                      & $AP_{3D}$   & \textbf{83.6} & \textbf{84.1} & \textbf{89.9} & \textbf{93.4} & 84.3          & \textbf{70.2} & 89.0          & \textbf{66.0} & \textbf{82.4} \\ 
\multicolumn{1}{c|}{}                                              & \multirow{2}{*}{0.5} & $AP_{BEV}$  & \textbf{84.2} & \textbf{82.7} & \textbf{91.6} & 96.1          & \textbf{85.5} & \textbf{72.2} & \textbf{94.6} & \textbf{67.9} & \textbf{84.7} \\
\multicolumn{1}{c|}{}                                              &                      & $AP_{3D}$   & \textbf{60.0} & \textbf{59.0} & \textbf{67.1} & 73.4          & 63.3          & \textbf{41.9} & \textbf{67.3} & \textbf{50.8} & \textbf{60.8} \\ \hline
\end{tabular}

}
  \label{tab:GT Sampling}%
\end{table*}

\begin{figure*}[ht]
  \centering
   \includegraphics[width=0.9\linewidth]{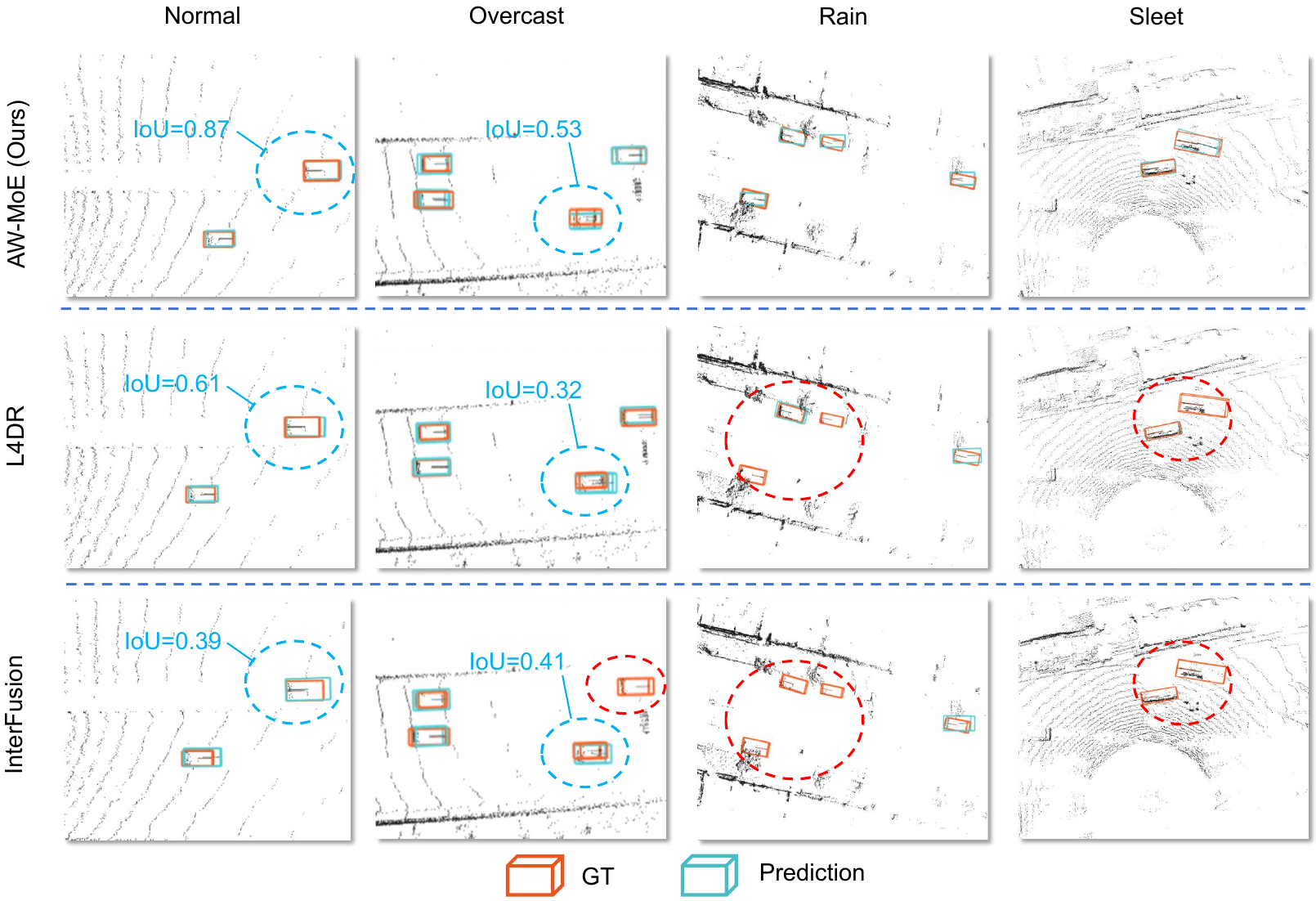}
   \caption{Comparison of our AW-MoE, L4DR~\cite{l4dr}, and InterFusion~\cite{interfusion} visualization results under Normal, Overcast, Rainy and Sleet weather conditions.}
   \label{fig:visual_1}
\end{figure*}

\begin{figure*}[ht]
  \centering
   \includegraphics[width=0.8\linewidth]{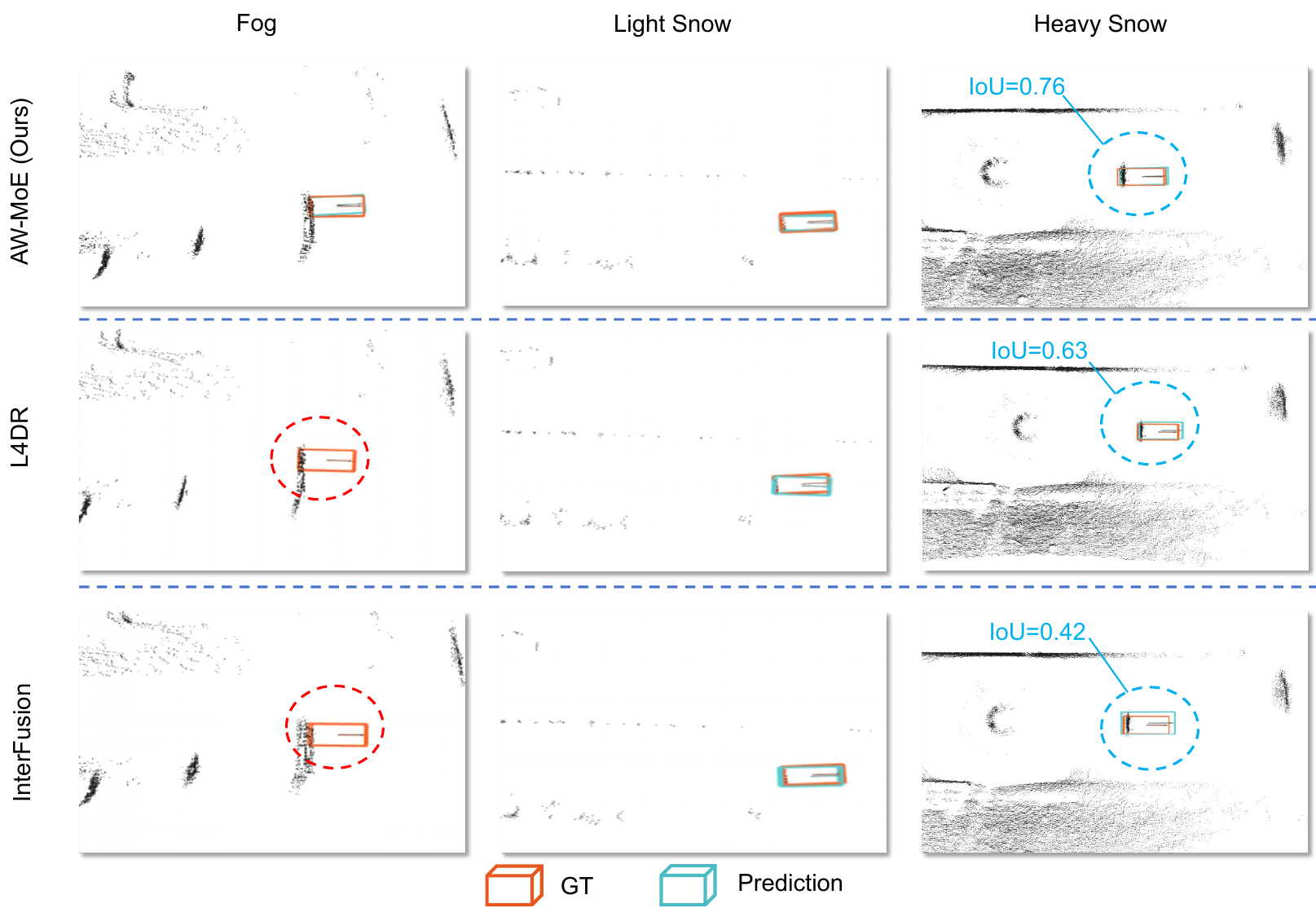}
\caption{Comparison of our AW-MoE, L4DR~\cite{l4dr}, and InterFusion\cite{interfusion} visualization results under Fog, Light Snow and Heavy Snow weather conditions.}
   \label{fig:visual_2}
\end{figure*}

\begin{table*}[ht]
\caption{Performance comparison of AW-MoE using different routing strategies: Point-cloud Feature-based Routing (PFR) and Image-guided Weather-aware Routing (IWR).}
\centering
\setlength{\tabcolsep}{9.5pt}
\resizebox{\textwidth}{!}{

\begin{tabular}{l|cccccccccc}
\hline
Routing               & IoU                  & Metric & Total         & Normal        & Overcast      & Fog           & Rain          & Sleet         & Light Snow    & Heavy Snow    \\ \hline
\multirow{4}{*}{PFR} & \multirow{2}{*}{0.3} & $AP_{BEV}$  & 58.7          & 87.3          & 94.3          & 23.1          & 77.7          & 34.7          & 70.2          & 70.9          \\
 &                      & $AP_{3D}$   & 52.8          & 84.0          & 89.3          & 11.4          & 73.0          & 29.1          & 64.8          & 63.6          \\
 & \multirow{2}{*}{0.5} & $AP_{BEV}$  & 55.6          & 82.1          & 91.4          & 22.9          & 74.4          & 31.4          & 69.6          & 66.9          \\
 &                      & $AP_{3D}$   & 35.4          & 57.8          & 66.5          & 6.2           & 52.4          & 16.6          & 46.6          & 49.9          \\ \hline
\multirow{4}{*}{IWR} & \multirow{2}{*}{0.3} & $AP_{BEV}$  & \textbf{88.2} & \textbf{87.7} & \textbf{94.5} & \textbf{96.7} & \textbf{88.8} & \textbf{81.0} & \textbf{95.4} & \textbf{70.2} \\
 &                      & $AP_{3D}$   & \textbf{83.9} & \textbf{84.2} & \textbf{90.0} & \textbf{95.3} & \textbf{84.4} & \textbf{72.9} & \textbf{90.2} & \textbf{64.0} \\
 & \multirow{2}{*}{0.5} & $AP_{BEV}$ & \textbf{84.2} & \textbf{82.8} & \textbf{91.6} & \textbf{96.3} & \textbf{85.3} & \textbf{75.0} & \textbf{94.7} & \textbf{66.4} \\
 &                      & $AP_{3D}$   & \textbf{61.5} & \textbf{59.0} & \textbf{67.2} & \textbf{85.7} & \textbf{63.5} & \textbf{43.3} & \textbf{70.1} & \textbf{53.1} \\ \hline
\end{tabular}

}
  \label{tab:Routing-method}%
\end{table*}

\begin{table*}[ht]
\caption{Performance ($AP_{3D}$) comparison between AW-MoE training strategy and direct end-to-end training. Non-normal denotes the aggregate of all non-normal weather conditions.}
\centering
\setlength{\tabcolsep}{6pt}
\resizebox{0.95\textwidth}{!}{

\begin{tabular}{l|cccccccccc}
\hline
Training Strategy                         & IoU & Total         & Normal        & Overcast      & Fog           & Rain          & Sleet         & Light Snow    & Heavy Snow    & Non-normal      \\ \hline
\multirow{2}{*}{Direct Training}          & 0.3 & 74.9          & 80.2          & 84.1          & 86.5          & 75.9          & 53.2          & 75.8          & 53.8          & 72.8          \\
  & 0.5 & 54.2          & 54.5          & 63.3          & 64.3          & 53.5          & 38.8          & 58.8          & 41.1          & 52.3          \\ \hline
\multirow{2}{*}{AW-MoE Training Strategy} & 0.3 & \textbf{83.9} & \textbf{84.2} & \textbf{90.0} & \textbf{95.3} & \textbf{84.4} & \textbf{72.9} & \textbf{90.2} & \textbf{64.0} & \textbf{83.9} \\
  & 0.5 & \textbf{61.5} & \textbf{59.0} & \textbf{67.2} & \textbf{85.7} & \textbf{63.5} & \textbf{43.3} & \textbf{70.1} & \textbf{53.1} & \textbf{61.5} \\ \hline
\end{tabular}

}
  \label{tab:Pretrained}%
\end{table*}

\begin{table*}[ht]
\caption{Effects of different top-K values in Image-guided Weather-aware Routing (IWR). Non-normal denotes the aggregate of all non-normal weather conditions.}
\centering
\setlength{\tabcolsep}{8.5pt}
\resizebox{0.95\textwidth}{!}{

\begin{tabular}{c|cccccccccc}
\hline
Parameter K            & IoU & Total         & Normal        & Overcast & Fog           & Rain & Sleet & Light Snow    & Heavy Snow & Non-normal      \\ \hline
\multirow{2}{*}{K = 1} & 0.3 & 83.9          & 84.2          & 90.0     & \textbf{95.3} & 84.4 & 72.9  & \textbf{90.2} & 64.0       & \textbf{83.9} \\
& 0.5 & \textbf{61.5} & 59.0          & 67.2     & \textbf{85.7} & 63.5 & 43.3  & \textbf{70.1} & 53.1       & 61.5          \\ \hline
\multirow{2}{*}{K = 2} & 0.3 & \textbf{84.2} & \textbf{84.9} & 90.0     & 95.0          & 84.8 & 72.8  & 88.4          & 64.8       & 82.8          \\
& 0.5 & 61.1          & \textbf{59.3} & 67.2     & 81.5          & 63.7 & 44.2  & 67.8          & 52.7       & \textbf{62.5} \\ \hline
\multirow{2}{*}{K = 3} & 0.3 & 83.2          & 84.1          & 89.0     & 94.1          & 84.7 & 69.9  & 88.2          & 64.4       & 82.2          \\
& 0.5 & 61.0          & 59.3          & 69.0     & 79.7          & 63.5 & 42.7  & 70.2          & 53.1       & 62.3          \\ \hline
\end{tabular}

}
  \label{tab:ToP K}%
\end{table*}

\begin{table}[t]
\caption{Robustness analysis of IWR under ambiguous weather conditions with varying Top-K values. (0.3 / 0.5) indicates the IoU value.}
\centering
\setlength{\tabcolsep}{5pt}
\resizebox{\columnwidth}{!}{

\begin{tabular}{c|c|c|c}
\hline
Parameter K            & Metric & Total (0.3 / 0.5)     & Non-normal (0.3 / 0.5) \\ \hline
\multirow{2}{*}{K = 1} & $AP_{BEV}$  &  82.8 / 79.1   &    86.0 / 82.0                  \\
& $AP_{3D}$   & 77.0 / 53.3           & 80.1 / 56.0          \\ \hline
\multirow{2}{*}{K = 2} & $AP_{BEV}$  &  \textbf{84.3 / 79.9} &          \textbf{87.8 / 82.7}    \\
& $AP_{3D}$   & \textbf{77.3  / 55.8} & \textbf{80.5 / 58.3} \\ \hline
\end{tabular}

}
  \label{tab:ambiguous}%
\end{table}

\subsection{Ablation Study}

\subsubsection{Effectiveness Analysis of Weather-Specific GT Sampling}
In this section, we compare the proposed Weather-Specific GT Sampling (WSGTS) with traditional Weather-Agnostic GT Sampling (WAGTS). As shown in Table~\ref{tab:GT Sampling}, WSGTS consistently outperforms WAGTS under both normal and adverse weather conditions. This improvement stems from WSGTS sampling ground-truth data exclusively from scenes matching the current weather, which avoids the insertion of mismatched GT that could compromise scene authenticity while still enabling effective data augmentation.

\subsubsection{Ablation on Expert Routing}
Table~\ref{tab:Routing} compares the weather classification capabilities of Point-cloud Feature-based Routing (PFR) and Image-guided Weather-aware Routing (IWR). IWR achieves approximately 99\% accuracy across all weather categories. In contrast, PFR struggles significantly in conditions like overcast, fog, and snow due to the inherent limitations of point clouds in capturing weather semantics. Furthermore, Table~\ref{tab:Routing-method} presents an ablation study on detection performance when integrating these routing methods into the MoE framework. IWR consistently achieves much higher detection accuracy than PFR across all weather conditions. This superior performance stems directly from IWR's ability to accurately classify the weather and route features to the optimal expert module. Conversely, PFR's poor routing accuracy severely degrades final detection performance. Together, these evaluations validate the effectiveness and ingenuity of the IWR design.

\subsubsection{Analysis of AW-MoE Training Strategy}
In Table~\ref{tab:Pretrained}, we compare our AW-MoE training strategy with direct training. The results demonstrate that our strategy significantly improves detection performance, particularly under adverse weather conditions. For example, under fog, $AP_{3D}$ at IoU=0.5 increases by 21.4\%. This improvement stems from pre-training each Weather-Specific Expert (WSE) using all-weather data, allowing the WSEs to acquire basic 3D object detection capabilities before further fine-tuning within AW-MoE, effectively mitigating the challenges posed by limited adverse-weather data.

\subsubsection{Parameter K in Image-guided Weather-aware Routing}
We conducted an ablation study on the parameter $K$ in the IWR module, with results shown in Table~\ref{tab:ToP K}. Overall, $K=1$ and $K=2$ yield similar performance, and both outperform $K=3$. To investigate the minimal difference between $K=1$ and $K=2$, we evaluated their performance under ambiguous weather conditions, where IWR is prone to misclassification (Table~\ref{tab:ambiguous}). In these scenarios, $K=2$ performs better than $K=1$. Routing to multiple experts mitigates the impact of classification errors, thereby enhancing robustness. However, this advantage is negligible in the overall metrics (Table~\ref{tab:ToP K}). Because IWR achieves approximately 99\% classification accuracy, misclassified cases are too infrequent to significantly affect global performance. Consequently, to achieve an optimal balance between computational efficiency and detection accuracy, we set $K=1$.

\subsection{Visualization Comparison}
To provide a more intuitive understanding, we visually compare our AW-MoE against the L4DR~\cite{l4dr} and InterFusion~\cite{interfusion} baselines across various weather conditions (Fig.~\ref{fig:visual_1} and Fig.~\ref{fig:visual_2}). The visualizations demonstrate two key improvements. First, AW-MoE effectively reduces missed detections caused by adverse weather (highlighted by red circles). Second, it regresses higher-quality 3D bounding boxes that align more closely with the ground truth (GT) (highlighted by blue circles). These enhancements stem from the AW-MoE design, which successfully mitigates the distribution discrepancy across different weather conditions.

\section{Conclusion}
\label{sec:Conclusion}

In this paper, we propose AW-MoE, the first framework to incorporate Mixture of Experts (MoE) for 3D detection under adverse weather, effectively addressing performance conflicts caused by large inter-weather data discrepancies in single-branch detectors. Specifically, the proposed Image-guided Weather-aware Routing (IWR) leverages the distinct visual characteristics of camera images to classify weather conditions. This ensures precise data routing to the optimal expert model, effectively overcoming the inherent limitations of point-cloud-based routing. Extensive experiments on the K-Radar dataset demonstrate the superiority and strong generalizability of AW-MoE. Overall, AW-MoE provides an effective framework for 3D object detection under adverse weather, enabling various detection algorithms to achieve optimal performance across different conditions, while incurring minimal impact on inference speed and computational cost. 

\bibliographystyle{IEEEtran}
\bibliography{ref}

\newpage

 




\vfill

\end{document}